\documentclass[conference]{IEEEtran}
\IEEEoverridecommandlockouts

\usepackage{cite}
\usepackage{amsmath,amssymb,amsfonts}
\usepackage{algorithmic}
\usepackage{graphicx}
\usepackage{textcomp}
\usepackage{xcolor}
\usepackage{amsthm,amssymb}
\usepackage[ruled,linesnumbered]{algorithm2e}
\usepackage{wasysym}
\usepackage{mathtools}
\usepackage{algorithmic}
\usepackage{subcaption}
\usepackage{wrapfig}
\usepackage{booktabs}
\usepackage{floatrow}
\usepackage{comment}
\usepackage{tabu}  
\usepackage{enumitem}
\usepackage{xspace}
\usepackage{flexisym}
\usepackage{float}
\floatstyle{plaintop}
\restylefloat{table}
\usepackage{colortbl}
\usepackage[utf8]{inputenc}
\usepackage{graphicx}
\usepackage{xcolor}
\usepackage{amsmath} 
\usepackage{diffcoeff} 
\usepackage{amsmath}   
\usepackage{amssymb}   
\usepackage{mathtools} 

\usepackage{enumitem}
\usepackage{caption}   
\def\BibTeX{{\rm B\kern-.05em{\sc i\kern-.025em b}\kern-.08em
    T\kern-.1667em\lower.7ex\hbox{E}\kern-.125emX}}

\title{Prompt-Guided Dual Latent Steering for Inversion Problems}

\author{Yichen Wu\\
Nanjing University of Posts and Telecommunications\\
China, Jiangsu, Nanjing\\
{\tt\small 	p22000114@njupt.edu.cn}
\and
Xu Liu\\
University of Washington\\
1410 NE Campus Pkwy, Seattle, WA\\
{\tt\small xliu28@uw.edu}
\and
Chenxuan Zhao\\
University of Alberta\\
116 St 85 Ave, Edmonton, Canada\\
{\tt\small 	chenxua2@ualberta.ca}
\and
Xinyu Wu\\
University of Washington\\
1410 NE Campus Pkwy, Seattle, WA\\
{\tt\small 	xinyuw23@uw.edu}
}

\begin{document}

\twocolumn[{
\renewcommand\twocolumn[1][]{#1}
\maketitle
\begin{center}
    \includegraphics[width=\textwidth]{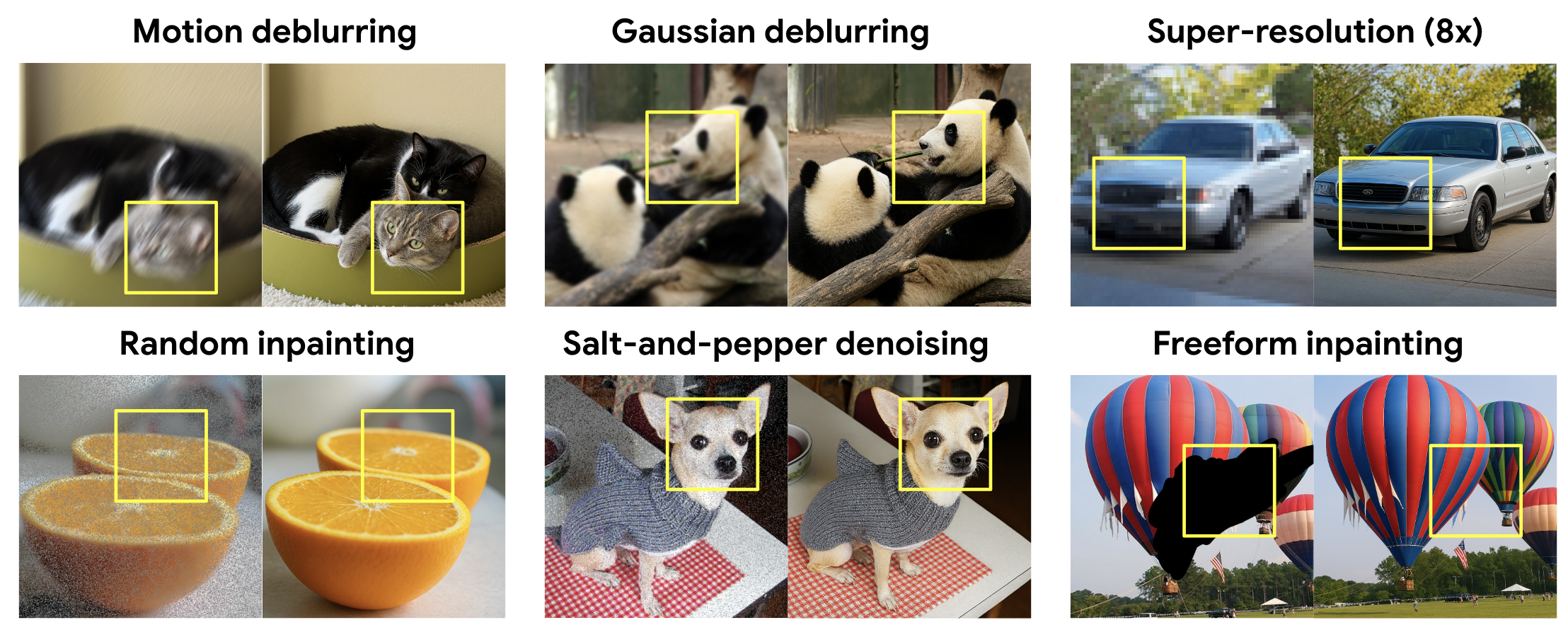}
    
    \captionof{figure}{
    Our proposed method PDLS serves as an inverse problem solver for various tasks and achieves high quality results.
    }
    \label{fig:main}
    
\end{center}
}]

\begin{abstract}
 Inverting corrupted images into the latent space of diffusion models is challenging. Current methods, which encode an image into a single latent vector, struggle to balance structural fidelity with semantic accuracy, leading to reconstructions with semantic drift, such as blurred details or incorrect attributes. 
To overcome this, we introduce Prompt-Guided Dual Latent Steering (PDLS), a novel, training-free framework built upon Rectified Flow models for their stable inversion paths. PDLS decomposes the inversion process into two complementary streams: a structural path to preserve source integrity and a semantic path guided by a prompt. We formulate this dual guidance as an optimal control problem and derive a closed-form solution via a Linear Quadratic Regulator (LQR). This controller dynamically steers the generative trajectory at each step, preventing semantic drift while ensuring the preservation of fine detail without costly, per-image optimization.
Extensive experiments on FFHQ-1K and ImageNet-1K under various inversion tasks, including Gaussian deblurring, motion deblurring, super-resolution and freeform inpainting,
demonstrate that PDLS produces reconstructions that are both more faithful to the original image and better aligned with the semantic information than single-latent baselines.

\end{abstract}


\section{Introduction}
With generative models prevailing across numerous tasks~\cite{BlackForestLabs2024,yang2024cogvideox,wang2025seal}, inversion within the generative models plays a crucial role. It seeks a latent representation that reproduces a given degraded or real‐world image and thereby enables accurate reconstruction and downstream editing. Formally, we observe a noisy signal $\mathbf{y}$:
\begin{equation}
\label{eq:forward}
\mathbf{y}=A\mathbf{x}+\mathbf{n},\qquad
\mathbf{n}\sim\mathcal{N}\!\bigl(\mathbf{0},\sigma_{y}^{2}I_{k}\bigr),
\nonumber
\end{equation}
where $A$ refers to a measurement operator~\cite{rout2024beyond}. The target is to sample from the Bayesian posterior $p(\mathbf{x}\mid\mathbf{y})$ or, in the diffusion formulation, from the continuum of posteriors $p_{t}(\mathbf{x}_{t}\mid\mathbf{y})$ along the reverse trajectory \cite{song2021ddim}.  Diffusion priors have achieved state‐of‐the‐art performance on super‐resolution, inpainting, and deblurring \cite{lugmayr2022sr3, saharia2022imagen, chung2022comecloser}, yet conventional inversion pipelines still rely on a single latent code obtained under a generic Gaussian noise assumption.  This mismatch leads to (i) poorly structured latents when the real degradation deviates from Gaussian white noise and (ii) reconstructions that ignore any semantic cues available at inference time.

\begin{figure*}
    \includegraphics[width=\linewidth]{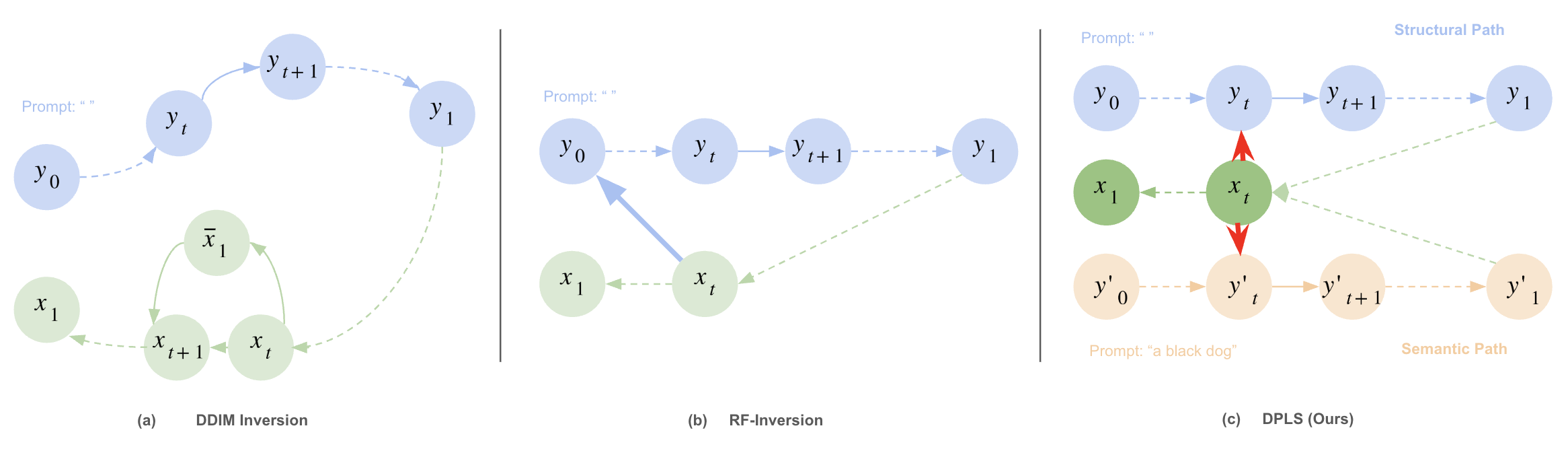}
    
    \caption{Model architecture overview and comparison with previous inversion-based methods.}
    
    \label{fig:exp:overview}
    \vspace{-0.1in}
\end{figure*}

Current diffusion models~\cite{rombach2022ldm,kawar2023imagic,mokady2023null,tumanyan2023plug} often employ a deterministic inversion process, such as DDIM inversion (Fig.\ref{fig:exp:overview}(a)), to map a real source image $y_0$ to a corresponding latent representation $y_1$. From this latent, a pretrained denoising model then reconstructs the image. While this approach can produce visually plausible results, its guidance relies predominantly on pixel-level consistency, which can lead to a semantic drift from the original input. This drift manifests as artifacts like localized blurring, loss of fine detail, or misalignment between the generated content and the intended semantics. Existing methods attempt to mitigate this by incorporating textual guidance or optimizing during inversion. However, these solutions are often computationally prohibitive, requiring either the optimization of numerous image-specific parameters~\cite{mokady2022nulltext} or the extensive test time fine-tuning of the diffusion model's architecture~\cite{rout2024beyond}, both of which demand significant computational and memory resources.

Another line of work uses Rectified Flow models during inversion~\cite{rout2024semantic} as shown in Fig.\ref{fig:exp:overview}(b). Rectified flow models learn straighter transport paths between noise and data distributions, resulting in simpler, near-linear ODE~\cite{chen2018neural} trajectories. This characteristic makes the inversion process more numerically stable and accurate, leading to high-fidelity reconstructions or editing with significantly less drift than traditional diffusion-based methods. However, the high-fidelity guarantee of standard RF inversion is largely predicated on the input image being a clean, on-manifold sample. When presented with a corrupted image (e.g., gaussian blur or motion blur), the inversion process is challenged. Forcing this corrupted sample through the straight-path inversion process can lead to a final latent variable that fails to accurately capture the original semantic information of the uncorrupted source. Consequently, while the subsequent reconstruction might be sharp and visually plausible, it often suffers from a loss of semantic fidelity. For example, the objects may be distorted or semantically incorrect (Fig.\ref{fig:quantitative_comparison}).

To address these limitations, we introduce Prompt-Guided Dual Latent Steering (PDLS), a novel, training-free framework that guides the reconstruction process along two complementary inversion paths: a structural path and a semantic path (Fig.\ref{fig:exp:overview}(c)). The structural path leverages the model's self-attention mechanisms with a null prompt to preserve the fine-grained geometry and appearance of the source image. Concurrently, the semantic path utilizes a user-provided text prompt to inject semantic priors, steering the reconstruction to align with the desired conceptual meaning.

Our method is built upon Rectified Flow (RF) models~\cite{liu2022flow}, which are distinguished by their straighter and more stable ODE trajectories between source and target distributions, enabling more faithful and predictable inversion. Within both inversion paths, we first apply a controlled forward ODE process, inspired by RF-Inversion~\cite{rout2024semantic}, to map the initial latent representation from a potentially corrupted, out-of-distribution space into the model's learned manifold. This crucial step helps preserve the fidelity of the source image throughout the generation process. During the reverse (generative) trajectory, we formulate the guidance mechanism as an optimal control problem. The objective is to find a controlled ODE that dynamically balances the influence of the structural and semantic paths. We derive a closed-form solution to this problem using a Linear Quadratic Regulator (LQR). The resulting optimal controller steers the latent state at each step, ensuring the final output harmoniously integrates the structural integrity of the source image with the semantic guidance of the text prompt.

Experimental results on both FFHQ-1K and ImageNet-1K with various corrupted image inputs demonstrate that PDLS generates reconstructions that exhibit superior structural fidelity and semantic alignment with the guiding prompt, outperforming existing state-of-the-art methods.
In summary, our contributions are as follows:
\begin{itemize}
\item We introduce a dual‐path inversion scheme that unifies structural fidelity and semantic precision while remaining model‐agnostic and training-free.
\item We derive a closed-form latent steering update, with its complexity matching the original denoising process, enabling semantic alignment at negligible extra cost.
\item We demonstrate consistent improvements on different inverse problems on ImageNet-1K and FFHQ-1K datasets, establishing PDLS as a plug-and-play method for existing latent diffusion models.
\end{itemize}

\section{Related Work} \label{sec:related}

Research on real or corrupted image manipulation with diffusion models can be mainly grouped into two work streams:  
(i) inversion-based methods and
(ii) prompt-guided latent control.
Our framework sits at the intersection of these two directions.

\subsection{Inversion-Based Methods} \label{sec:related:inv}
The deterministic ODE introduced by DDIM inversion~\cite{song2021ddim}
maps a real image to its latent origin in a single pass, but it
implicitly assumes the degradation matches the original forward schedule,
often mis-aligning high-frequency details.  
DiffPIR~\cite{zhu2023denoising} and other plug-and-play variants~\cite{zhu2023denoising,martin2024pnp} inject
classical priors into the sampling loop yet still rely on a single
inversion path and ignore textual semantics.  
Null-Text Inversion~\cite{mokady2022nulltext} optimizes a special “null”
token per image, which improves fidelity at the cost of dozens of
gradient steps; 
Most related to our work, Dual-Conditional Inversion
(DCI)~\cite{ren2025dci} jointly refines the latent with a reference image
and a prompt using a fixed-point scheme; however, it requires an
inner-loop solver for every test image.

RF-Inversion \cite{rout2024semantic} formulates image inversion as a dynamic optimal control problem for the rectified-flow ODE. By deriving the controlled ODE via a linear-quadratic regulator, it obtains a closed-form vector field that is provably equivalent to a rectified stochastic differential equation, so a single forward integration, performed in one pass and without learnable parameters, recovers the latent origin of a real or degraded image. 
However, the method still operates along a single inverse path and therefore does not consider semantic cues. In contrast, we compute two inversion latents and fuse them on the fly through a lightweight, time-decaying steering rule, removing iterative optimization while fully exploiting multi-modal guidance.

\begin{figure*}[th]
    \centering
    \includegraphics[width=0.85\linewidth]{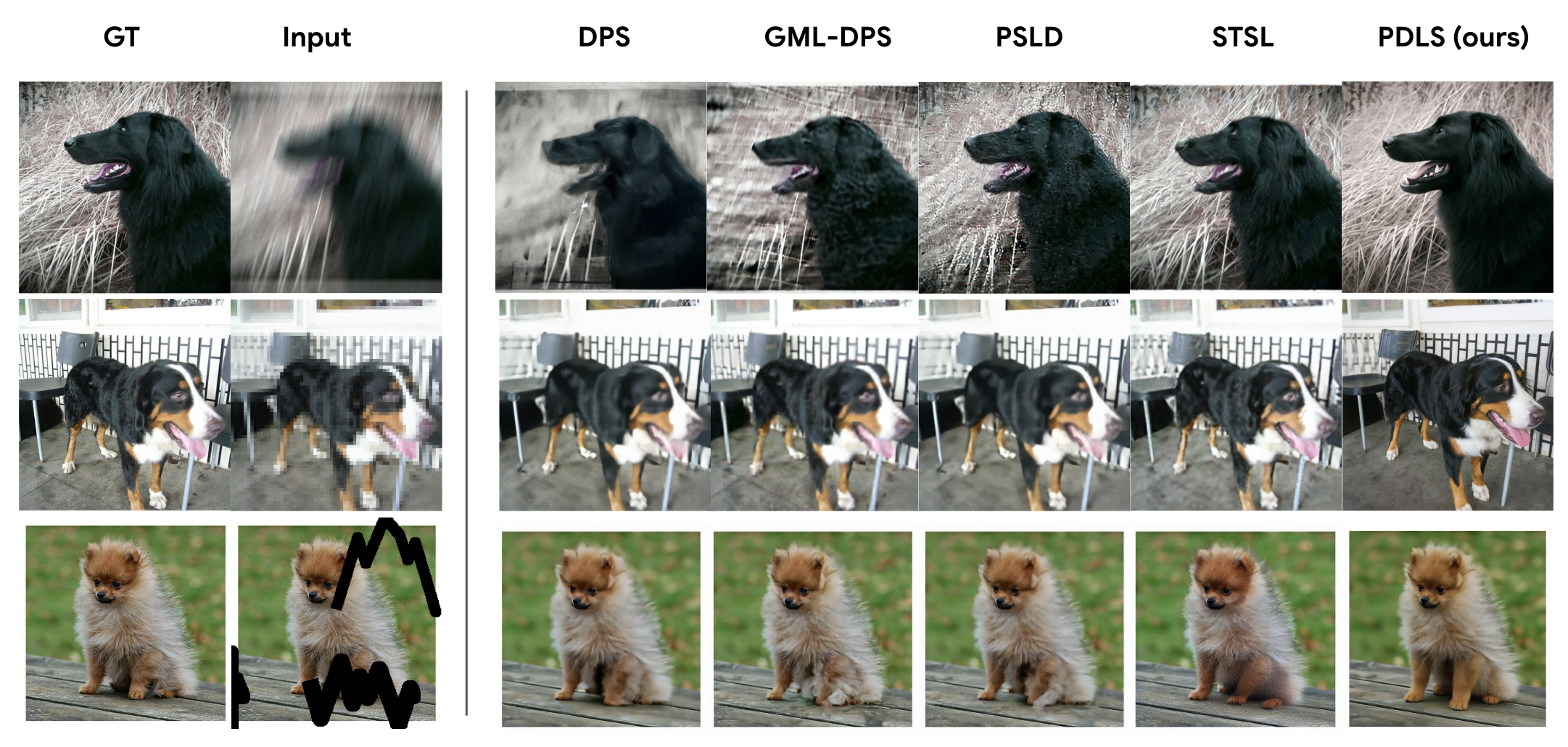}
    \caption{Comparison of PSNR and LPIPS between our method and baseline inversion/restoration methods across 100 samples. Our approach consistently achieves higher perceptual quality and fidelity.}
    \label{fig:quantitative_comparison}
\end{figure*}

\subsection{Prompt-Guided Latent Control}
\label{sec:related:prompt}
Text prompts have become the de-facto interface for diffusion generators.
DiffEdit~\cite{couairon2022diffedit} replaces semantics inside a user
mask via latent resampling, whereas InstructPix2Pix
\cite{brooks2023instructpix2pix} fine-tunes a generator on instruction–image pairs to enable multi-step edits. Prompt-to-Prompt (P2P)~\cite{hertz2022p2p} controls
cross-attention for local editing but likewise hinges on a single latent code.  
DreamBooth~\cite{ruiz2022dreambooth}, Textual
Inversion~\cite{gal2022textualinv} and RB-Modulation~\cite{rout2024rb} personalize a model with only a few or only one reference
image, but these methods target content synthesis and pay little
attention to structural fidelity.  
For restoration, Diff-Restorer~\cite{zhang2024diff} injects semantic
and degradation embeddings into the U-Net, while PromptDiff
\cite{zhang2024diff} jointly optimizes pixel and CLIP losses~\cite{radford2021learning}.  Both
techniques modify the original network or demand extra
training.  
More recently, ControlNet~\cite{zhang2023controlnet} and T2I-Adapter
\cite{liu2023t2iadapter} append learnable branches that feed edge, depth,
or pose maps into diffusion layers, achieving precise spatial control at
the cost of additional parameters.  
Latent blending methods~\cite{karras2019style} perform a
static linear interpolation between two codes, which fails to
enforce consistency along the full sampling trajectory.  

Our method keeps the pretrained model frozen and manipulates
only the latent variables, preserving structure without any network
surgery.
Besides, our work can be viewed as a trajectory-aware, zero-overhead alternative:
we introduce a time-varying interpolation between two inversion
paths, steering every denoising step without changing network weights.

\section{Method}\label{sec:method}

\subsection{Preliminary}\label{sec:method:background}
Throughout this paper, we adopt the rectified-flow model Flux~\cite{dockhorn2021score} as backbone and modify our inversion method based on the RF-Inversion~\cite{rout2024semantic}.  
This provides a geometry-faithful latents on which we subsequently attach semantic guidance, yielding a dual-path model that balances realism and semantic fidelity. The rectified-flow models are trained by solving
\begin{equation}
\frac{d\mathbf{x}_t}{dt} = \mathbf{v}_\theta(\mathbf{x}_t,t), 
\qquad t\in[0,1],\;
\mathbf{x}_{0}\sim\mathcal{N}(0,I),
\label{eq:rf_ode}
\end{equation}
where the learned vector field $\mathbf{v}_\theta$ deterministically transports the sample from isotropic Gaussian noise at $t=0$ to the data manifold at $t=1$.

Unlike diffusion models that learn a stochastic SDE~\cite{song2020score,vahdat2021score,lipman2022flow}, Rectified Flow models learn a deterministic ODE with its velocity field carries every point straight from the noise prior to the data manifold \cite{liu2022flow}.  
The resulting sampler generates high-quality images in only a few Euler steps because each trajectory is a near-linear “shortcut’’ through latent space.

The recent proposed inversion method for Rectified Flow models, RF-Inversion~\cite{rout2024semantic}, shows that the same rectified ODE can be inverted efficiently by solving a dual-objective optimal-control problem. 
The key outcome is a rectified vector field 
\begin{equation}
\mathrm{d}Y_t
  =\bigl[u_t(Y_t)+\gamma\bigl(u_t(Y_t\mid\mathbf{y}_1)-u_t(Y_t)\bigr)\bigr]\,\mathrm{d}t
\label{eq:rf_inv_ode}
\end{equation} that gradually pulls atypical samples back toward the high-density manifold. 
$\mathbf{y}_1$ refers to the original image and $u_t(Y_t\mid\mathbf{y}_1)$ is calculated by solving a LQR problem.
Intuitively, the field acts as an adaptive “gravity’’ where the points far from typical data receive stronger correction, so the reconstructed image becomes both realistic and editable without per-image optimization.

\subsection{Dual Inversion Paths}
\label{sec:dual_anchors}
Deterministic latent inversions such as DDIM~\cite{song2021ddim,li2024pre} reconstruct an input image by running the forward noising ODE and then reversing it.  
Because the forward phase discards high-frequency detail, the latent trajectory drifts away from the data manifold, producing blur and color bleeding, a phenomenon known as semantic drift.  
RF-Inversion~\cite{rout2024semantic} mitigates this drift with a rectified flow, recovering global geometry; however, we can see in Fig.\ref{fig:exp:ablation_1}, with structural path only inversion, there are still misalignment issues which are discussed in Section \ref{sec:Experiments:Ablation}.
As illustrated in Fig.\ref{fig:exp:overview}(c), PDLS mitigates this limitation by computing two independent inversion paths: structural path $Y_t$, obtained with a null prompt, which preserves pixel-level geometry yet conveys little semantic context. Semantic path $Y'_t$ ,computed with the user’s descriptive prompt which injects high-level priors but may hallucinate under severe corruption.
Both trajectories are produced once, in closed form with the RF-Inversion solver, and require no per-image optimization or network fine-tuning is required.  
The subsequent Dynamic Latent Steering module (Section~\ref{sec:latent_steer}) fuses these two paths during the reverse process, balancing structural fidelity against prompt consistency.  

\subsection{Prompt Informativeness}
\label{sec:prompt_inform}

For the semantic path, when the text prompt is too short, for instance only the noun “dog”, the prompt supplies little more than a class label. The model receives no guidance on pose, environment or style and must infer these factors autonomously, a situation that increases hallucination and produces large variance in color and geometry across runs. At the opposite extreme, an overly long prompt packed with stylistic adjectives and scene directives overwhelms the semantic path. The attention weights skew toward the prompt keys, the latent trajectory becomes dominated by high-level semantics, and fine structural cues from the structural path are partially ignored; we observe color shifts, exaggerated lighting effects and occasional divergence of the reverse process.

A series of ablations therefore searched for a middle ground. The most effective recipe is a high-resolution reference image paired with a concise descriptive phrase composed of one noun and one or two factual modifiers, for example “green sea turtle under daylight”.  In this configuration, the structural path conveys high-level information such as pose, texture, and lighting, while the short prompt fixes class identity and suppresses drift toward visually similar species.

\subsection{Dynamic Latent Steering}
\label{sec:latent_steer}

In the reverse process, we aim to progressively reconstruct high-quality image from the previously calculated dual inversion paths, which leverage both structural fidelity and semantic information. To achieve this, we initialize the reverse diffusion process with the latent $y_1$ from the structural inverse path, which captures low-level content and structure from the corrupted input. All the intermediate latents during the inverse process have been stored so the memory overhead is minimal.

At each denoising step,
we formulate the latent steering as a controlled ODE.
Let \(X_t\) be the latent trajectory, and let \(y_t\) and \(y_t'\) be the
unguided and prompt-guided reference trajectories from structural and semantic inversion paths at step $t$.  
Define the averaged target as \(\bar y_t=\tfrac12\,(y_t+y_t')\).
With guidance strength \(\eta\in[0,1]\),
\begin{equation}
\label{eq:control_ode}
\mathrm dX_t
  =\bigl[\,v_t(X_t)+\eta\bigl(v_t(X_t\!\mid\!\bar y_t)-v_t(X_t)\bigr)\bigr]\,
   \mathrm dt,\quad X_0=y_1,
\end{equation}
where \(v_t(\cdot)\) is the reverse SDE drift.
Solving the modified LQR
\begin{multline}\label{eq:value}
V(c)=\int_{0}^{1}\tfrac12\lVert c(Z_t,t)\rVert_2^{2}\,\mathrm dt
      +\tfrac{\lambda}{2}\lVert Z_1-\bar y_t\rVert_2^{2}, \\
\mathrm dZ_t=c(Z_t,t)\,\mathrm dt,\quad Z_0=y_1.
\end{multline}
gives the optimal control
\begin{equation}
c(Z_t,t)=\frac{\bar y_t-Z_t}{1-t}.
\end{equation}
One Euler step of this ODE recovers, revealing the
update as a time-varying optimal controller that monotonically decreases
\(\lVert X_t-\bar y_t\rVert_2\).

\begin{table*}[!thb]
\centering
\setlength{\tabcolsep}{2pt}
\resizebox{0.8\textwidth}{!}{
\begin{tabular}{l|ccc|ccc|ccc|ccc}
\toprule
{} & \multicolumn{3}{c|}{\textbf{SR ($\times 8$)}} &
     \multicolumn{3}{c|}{\textbf{Motion Deblur}} &
     \multicolumn{3}{c|}{\textbf{Gaussian Deblur}} &
     \multicolumn{3}{c}{\textbf{Freeform Inpainting}} \\
\cmidrule(lr){2-13}
\textbf{Method} &
LPIPS$\downarrow$ & PSNR$\uparrow$ & SSIM$\uparrow$ &
LPIPS$\downarrow$ & PSNR$\uparrow$ & SSIM$\uparrow$ &
LPIPS$\downarrow$ & PSNR$\uparrow$ & SSIM$\uparrow$ &
LPIPS$\downarrow$ & PSNR$\uparrow$ & SSIM$\uparrow$ \\
\midrule
DPS~\cite{chung2022diffusion}          & 0.529 & 29.95 & 71.9 & 0.559 & 28.98 & 71.8 & 0.699 & 28.34 & 56.2 & 0.513 & 28.88 & 72.0 \\
DiffPIR~\cite{zhu2023denoising}
                        & 0.790 & 28.15 & 42.7 & 0.595 & 28.43 & 67.9 & 0.600 & 29.34 & 73.1 & 0.575 & 28.55 & 42.9 \\
\midrule
LDPS~\cite{rout2023solving}        & 0.340 & \underline{32.60} & \textbf{93.5} & 0.425 & 31.38 & 85.8 & 0.380 & 32.20 & 92.2 & 0.392 & \textbf{33.75} & \textbf{92.1} \\
GML-DPS~\cite{rout2023solving}     & 0.376 & 32.50 & 92.8 & 0.408 & 31.35 & 87.8 & 0.374 & \textbf{32.25} & 92.6 & 0.360 & 31.30 & 88.8 \\
PSLD~\cite{rout2023solving}        & 0.415 & 31.40 & 89.0 & 0.412 & 31.35 & 87.6 & 0.383 & 32.18 & 92.6 & 0.410 & 32.15 & 89.9 \\
STSL~\cite{rout2024beyond}      & \textbf{0.335} & 31.90 & 91.4 & \textbf{0.321} & \underline{31.71} & \underline{90.0} & \underline{0.308} & \underline{32.24} & \underline{94.1} & \textbf{0.311} & 31.19 & 90.7 \\
\cmidrule(l){1-13}
PDLS(ours)  & \underline{0.341} & \textbf{33.00} & \underline{93.1} &
                          \underline{0.352} & \textbf{31.92} & \textbf{90.6} &
                          \textbf{0.300} & \underline{32.24} & \textbf{94.2} &
                          \underline{0.313} & \underline{32.70} & \underline{92.0} \\
\bottomrule
\end{tabular}}
\caption{Quantitative results of different inversion tasks on FFHQ-1K. Best results are in \textbf{bold} and the second best are \underline{underlined}.}
\label{tab:ffhq-updated}
\end{table*}

\subsection{Time-decaying Steering Schedule}
\label{sec:time-schedule}
The influence of the steering controller on the main generative trajectory, $X_t$, should not be uniform throughout the reverse process. In the initial steps, when $X_t$ is highly noisy and far from the target manifold, a strong control is needed to guide it towards the target $\bar{y}_t$. In contrast, during the later steps, when $X_t$ already possesses fine-grained details, an overly aggressive control can be disruptive and misguide the generation to the corrupted input images, so a smaller control strength is required for subtle refinement.

To implement this dynamic control, we introduce a Time-decaying Steering Schedule. Instead of using a fixed controller strength $\eta$, we make it a function of the normalized timestep $t \in [0, 1]$. We employ a cosine decay schedule, which proves highly effective:
\begin{equation}
\label{eq:cosine_decay_schedule}
\eta(t) = \frac{\eta_{\max}}{2} \left(1 + \cos\left(\pi \cdot t\right)\right)
\end{equation}
Here, $\eta_{\max}$ is the initial guidance strength at $t=0$. This schedule applies the strongest control at the beginning of the process and gradually fades its influence to zero by the final step.

This schedule acts as a dynamic modulator for the optimal controller derived from the LQR problem (eq.~\ref{eq:control_ode}). While the optimal control $c(Z_t, t)$ naturally intensifies its correction via the $\frac{1}{1-t}$ term as the process unfolds, our decaying schedule $\eta(t)$ tempers this effect in the later stages. This ensures that the global structure is established early, while the final, delicate phase prioritizes the faithful reconstruction of details, leading to a high-quality output.

\section{Experiments}

\subsection{Experimental Setup}
\label{sec:exp:setup}

\noindent\textbf{Datasets and baselines.} 
We report our results on FFHQ-1K and the ImageNet-1K datasets. The images are resized to $512{\times}512$ and each dataset contains 1000 images identical to prior latent-diffusion works.
We compare our approach with state-of-the-art diffusion inverse problem solvers PSLD~\cite{rout2023solving}, STSL \cite{rout2024beyond}, LDPS~\cite{rout2023solving} and GML-DPS~\cite{rout2023solving}, as well as pixel-domain methods DPS~\cite{chung2022diffusion} and DiffPIR~\cite{zhu2023denoising}, using the official code and hyper-parameters.

\noindent\textbf{Tasks and metrics.} 
Four experiment settings are considered: motion deblurring produced by a $61{\times}61$ motion blur kernel with intensity 0.5; super-resolution from inputs reduced by factors of $8{\times}$; Gaussian deblurring generated with a $61{\times}61$ kernel of standard deviation $\sigma=3.0$; free-form inpainting with 10\

We follow \cite{rout2024beyond} on the evaluation criteria. 
Perceptual quality is assessed with LPIPS, where lower values indicate superior perceptual fidelity. Additionally, we employ two traditional image quality metrics: PSNR and SSIM. For both PSNR and SSIM, higher values indicate better image quality, reflecting a closer resemblance to the ground truth in terms of signal-to-noise ratio and structural integrity, respectively. These metrics collectively provide a comprehensive evaluation of the reconstruction and generation capabilities.

\noindent\textbf{Implementation details.}
We use a fixed controller strength 0.5 during inverse as used in RF-inversion’s controller forward ODE. Our method is built with Flux.1-dev with 28 denoising steps. We fix $\eta_{max}$ as 0.5 for all the experiments. All the other model specific hyper-parameters follow original configurations in the Flux model and RF-inversion. We use the noise level $\sigma_y = 0.01$ and image size 512$\times$512. All experiments run on a single NVIDIA A100 GPU.

\begin{table*}[!thb]
\centering
\setlength{\tabcolsep}{2pt}
\resizebox{0.8\textwidth}{!}{
\begin{tabular}{l|ccc|ccc|ccc|ccc}
\toprule
{} & \multicolumn{3}{c|}{\textbf{SR ($\times 8$)}} &
     \multicolumn{3}{c|}{\textbf{Motion Deblur}} &
     \multicolumn{3}{c|}{\textbf{Gaussian Deblur}} &
     \multicolumn{3}{c}{\textbf{Freeform Inpainting}} \\
\cmidrule(lr){2-13}
\textbf{Method} &
LPIPS$\downarrow$ & PSNR$\uparrow$ & SSIM$\uparrow$ &
LPIPS$\downarrow$ & PSNR$\uparrow$ & SSIM$\uparrow$ &
LPIPS$\downarrow$ & PSNR$\uparrow$ & SSIM$\uparrow$ &
LPIPS$\downarrow$ & PSNR$\uparrow$ & SSIM$\uparrow$ \\
\midrule
DPS~\cite{chung2022diffusion} & 0.538 & 29.15 & 72.9 & 0.556 & 28.98 & 71.9 & 0.627 & 28.43 & 56.9 & 0.689 & 27.73 & 56.8 \\
DiffPIR~\cite{zhu2023denoising} & 0.791 & 28.12 & 42.6 & 0.593 & 28.92 & 67.1 & 0.674 & 29.86 & 73.2 & 0.619 & 29.18 & 72.7 \\
\midrule
LDPS~\cite{rout2023solving} & 0.354 & \underline{32.54} & \textbf{93.5} & 0.433 & 31.21 & 86.3 & 0.392 & 31.99 & 91.8 & 0.387 & 32.25 & 91.7 \\
GML-DPS~\cite{rout2023solving} & 0.368 & 32.34 & \underline{92.7} & 0.408 & 31.43 & 88.0 & 0.360 & \textbf{32.33} & 93.1 & 0.363 & \underline{32.38} & \underline{92.8} \\
PSLD~\cite{rout2023solving} & 0.402 & 31.39 & 88.9 & 0.407 & 31.37 & 87.6 & 0.382 & 32.19 & 91.4 & 0.354 & 31.98 & \underline{92.8} \\
STSL~\cite{rout2024beyond} & \textbf{0.335} & 31.77 & 91.3 & \textbf{0.321} & \underline{31.71} & \underline{90.0} & \underline{0.316} & \underline{32.28} & \underline{94.1} & \underline{0.321} & 32.10 & \textbf{94.3} \\
\cmidrule(l){1-13}
PDLS(ours) & \underline{0.340} & \textbf{33.03} & 91.8 & \underline{0.349} & \textbf{32.16} & \textbf{91.02} & \textbf{0.308} & \underline{32.28} & \textbf{94.49} & \textbf{0.305} & \textbf{32.80} & 92.5 \\
\bottomrule
\end{tabular}}

\vspace{-1ex}
\caption{Quantitative results of different inversion tasks on ImageNet-1K. Best results are in \textbf{bold} and the second best are \underline{underlined}.
}
\label{tab:ffhq-imagenet-1k}
\vspace{-1ex}
\end{table*}

\subsection{Qualitative Results}

Fig.\ref{fig:quantitative_comparison} presents a side-by-side comparison of our PDLS against five recent baselines, DPS~\cite{chung2022diffusion}, GML-DPS~\cite{rout2023solving}, PSLD~\cite{rout2023solving}, and STSL~\cite{rout2024beyond}, under three degradation cases: motion blur, super-resolution (8x) and freeform inpainting. The leftmost image (``GT'') is the original ground-truth image, followed by the corrupted input and the outputs of each competing method.
In the first row we demonstrate motion deblurring. All baselines struggle to restore the dog’s eyes. DPS leaves ghosted pupils; PSLD produces cross-hatched artifacts; GML-DPS and STSL soften the entire orbit, yielding a glassy look.
PDLS reconstructs a crisp iris highlight and a well-defined eyelid edge while simultaneously removing streaks in the background reeds, indicating superior fine-structure recovery.

In the second example of super-resolution (8x), the competing methods still contain noticeable blur. For example, DPS and PSLD display checkerboard ringing, and even STSL leaves a hazy muzzle.
PDLS, starting its reverse path from an intermediate latent, inherits the correct global silhouette but re-injects high-frequency detail through dual-path guidance.
The restored nostril edges and tongue highlight match the GT far more closely than any baseline.

For free-form inpainting, when the broad black stroke occludes the puppy, every baseline removes the mark but mishandles geometry, most visibly the dog’s right fore-foot.
DPS leaves a warped outline, and STSL misaligns the paw pad.
PDLS synthesizes a natural, anatomically consistent foot that blends with the table texture and matches the pose seen in the GT.
Across all three scenarios the other methods either amplify texture noise or introduce prompt-driven geometric drift.
By fusing an unguided inversion (structural path) with a prompt-guided inversion (semantic path) at every step, PDLS keeps low-level fidelity and high-level consistency in balance, yielding sharper eyes, cleaner fur, and correctly shaped limbs where alternative methods fail.

\begin{figure*}
    \includegraphics[width=\linewidth]{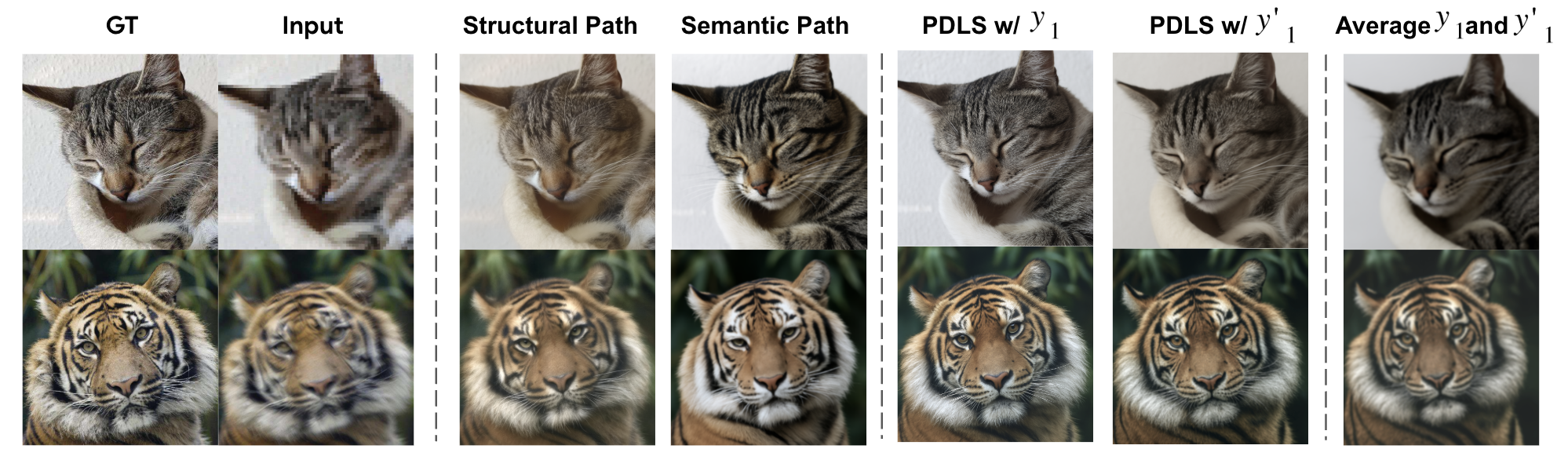}
    
\caption{Ablation study on various experiment settings. Column three and four compares results with different inverse paths. Column five to seven compares different initialized noise for the reverse process.
}
    
    \label{fig:exp:ablation_1}
    
\end{figure*}

\subsection{Quantitative Results}
\label{sec:quantitative}

Table~\ref{tab:ffhq-updated} shows that PDLS achieves the significant perceptual gain on the FFHQ–1K motion-deblur task: LPIPS stays close with the best baseline value of 0.321 (STSL) to 0.352, only 0.031 difference, while SSIM increases from 90 to 90.6, a 0.67\
For Gaussian deblurring, the best baseline(STSL) LPIPS value falls is 0.308 compare to 0.300(PDLS), a 2.6\
On free-form inpainting the best baseline method (STSL) attains an LPIPS of 0.311, However, our method (PDLS) is only 0.002 lower than it, while preserving a competitive SSIM of 92.0, indicating more faithful texture synthesis near irregular masks.

A similar pattern is observed on ImageNet-1K in Table~\ref{tab:ffhq-imagenet-1k}.  
For motion deblurring, PSNR rises from best baseline 31.71 (STSL) to 32.16, a 1.4\
Because ImageNet covers wide pose and lighting variation, these results confirm that the dual-path strategy generalizes beyond face-centric data.

There are three key factors that explain the improvements. First, the structural inversion path without a prompt provides accurate geometry and the semantic path with a concise prompt supplies class-level information. Combining them at every reverse step guides the latent toward a region where both priors agree, which lowers perceptual error. Second, the time-decaying steering schedule uses larger coefficients in early iterations to pull the trajectory away from degradation-induced minima and smaller coefficients later to preserve fine detail, which accounts for the strong SSIM increase on motion-blurred inputs. Finally, we employ the forward ODE as used in RF-Inversion that pushes atypical samples toward high-density regions, so prompt-aware steering compounds this rectification and yields reconstructions that remain realistic while better matching the user prompt.

\begin{figure*}[t]
  \centering
  \includegraphics[width=\linewidth]{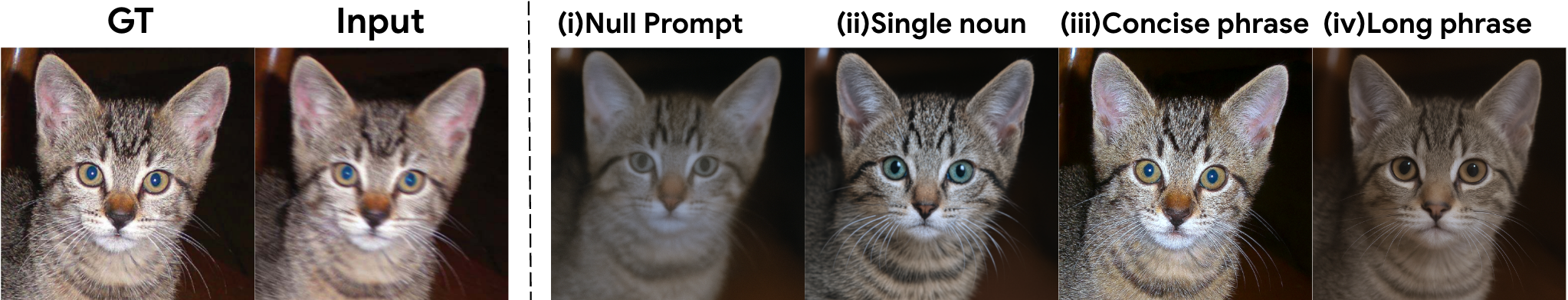}
  \caption{Results with different prompts used in the semantic path as illustrated in Section~\ref{sec:Experiments:Ablation}. Concise guidance (iii) restores the correct eye color and fur detail, whereas under (i)-(ii) or over-specified prompts (iv) blur or hallucinate attributes.}
  \label{fig:exp:ablation_2}
  
\end{figure*}

\subsection{Ablation Study}\label{sec:Experiments:Ablation}

\noindent \textbf{Effectiveness of dual inversion paths.}
Fig.\ref{fig:exp:ablation_1} compares three configurations: semantic path only, structural path only, and PDLS which contains both paths.  
The prompts used for the two examples are:  
“a sleeping cat on a beige couch” and “a close-up portrait of a Bengal tiger in dense jungle”.
Across all examples, our dual-path PDLS (initialized w/ $y_1$) yields the highest perceptual sharpness and semantic fidelity, confirming that prompt information complements the structural path.

In the first row, we show 8x super-resolution task with a sleeping cat image.  
The structural path retains the cat’s curled pose and soft side–lighting, yet because it is driven solely by the heavily down–sampled input it inherits strong low-pass blur. For example, its guard-hairs fuse into a fuzzy halo and the blockiness remains on the wall. The Semantic Path, conditioned only on the prompt, sharpens tabby stripes and reveals fine wall grain, but by over-relying on the model’s cat prior leads to a cooler color cast and slightly shifts the muzzle outline.

The second row of Fig.\ref{fig:exp:ablation_1} demonstrates motion deblurring. 
Structural inversion preserves the head contour but smears whiskers and blurs stripe boundaries, a direct consequence of pushing a blur‑corrupted image through a straight inversion path.
Semantic inversion, recovers vivid stripes and bright eye highlights but exaggerates contrast around the jaw, an over‑correction from the model’s tiger prior.
Our method reinstates high-contrast stripe boundaries and resolves whisker tips.

\noindent\textbf{Impact of different prompt guidance in the semantic path.}
Fig.\ref{fig:exp:ablation_2} fixes the same motion-blurred input while
varying only the prompt in the semantic path.
(i) Null prompt, with no textual cue the solver depends solely on the corrupted pixels. This is equal to structural path only, and we observed similar color distortion as seen in the previous ablation study.
The reconstruction regresses toward a blurry average: the
kitten’s blue–green irises fade to gray, individual guard-hairs are wiped out.
(ii) Single noun “Cat”, a class label supplies minimal semantics, so the model still borrows most of its guidance from the degraded image. Eye color drifts
toward an unsaturated amber and the tabby stripes soften, revealing that
a short prompt is insufficient to recover the semantic drift.
(iii) Concise phrase, the short factual description
“Close-up of a young tabby kitten, ears perked and bright
blue-green eyes” introduces just three concrete attributes. This limited yet specific
information is enough to reinforce the correct eye hue, restore crisp fur texture, and recover the true stripe frequency without overriding the structural path.
(iv) Long phrase, the extended sentence injects many extraneous adjectives, ``Wide-eyed tabby kitten with perked ears and long white whiskers; dark forehead stripes form an M while silvery-gray fur shifts to sandy tones under the chin against a dim wooden backdrop.'' The additional constraints pull the semantic path away from the visual evidence, hallucinating darker ``M-shaped'' forehead stripes and an overall cooler palette that is absent from the ground truth.

Under-specified prompts (i–ii) lack the detail needed to correct blur-induced loss, whereas an over-specified prompt (iv) can dominate the reverse process and introduce hallucinations.  A concise, factual
prompt (iii) best balances the dual-paths, yielding faithful color and fine-grained detail.

\noindent\textbf{Difference of initialization $y_1$ and $y'_1$.}
As shown in the last three columns in the Fig.~\ref{fig:exp:ablation_1}, initializing the reverse process from either the structural path ($y_1$) or the semantic path ($y'_1$) proves effective. This efficacy stems from the latent steering control algorithm's ability to dynamically adjust the reverse trajectories during the denoising iterations.

However, a noticeable degradation like blurriness occurs when $y_1$ and $y'_1$ are mixed (e.g., averaged). This phenomenon is attributed to the RF-inversion process, which shifts both $y_1$ and $y'_1$ towards a typical distribution via eq.~\ref{eq:rf_inv_ode}. Consequently, their combination deviates from this typical distribution, leading to the observed decline in results. For consistency and to ensure a fair comparison with other methods, we standardize our initialization to $y_1$ from the structural path in all experiments.

\section{Conclusion}
In this work we introduced Prompt-Guided Dual Latent Steering (PDLS), a training-free, rectified-flow framework for inverse image problems that simultaneously preserves structure and respects user semantic input. PDLS first traces two complementary inversion paths: a structural path, computed with a null prompt, that recovers low-level geometry, and a semantic path, guided by a concise text prompt, that restores high-level meaning. We cast their fusion as an optimal-control problem, solve it in closed form with a linear–quadratic regulator, and temper the resulting control with a cosine time-decay schedule. This lightweight controller adds virtually no runtime or parameters, yet consistently outperforms state-of-the-art diffusion and latent posterior samplers.
Future work will extend PDLS to video restoration, task-specific perceptual rewards, and multi-modal paths beyond text, enabling richer and more controllable inversion pipelines.

\bibliographystyle{IEEEtran}
\bibliography{references}

\end{document}